\pdfoutput=1

\documentclass[11pt]{article}

\usepackage[final]{acl}

\usepackage{times}
\usepackage{latexsym}

\usepackage[T1]{fontenc}

\usepackage[utf8]{inputenc}

\usepackage{microtype}

\usepackage{inconsolata}

\usepackage{graphicx}

\usepackage{longtable}

\usepackage{forest,adjustbox}
\useforestlibrary{linguistics}
\forestapplylibrarydefaults{linguistics}
\usepackage{avm}

\usepackage{booktabs}
\usepackage{multicol}
\usepackage{multirow}
\usepackage{subcaption}

\title{Comparing LLM-generated and human-authored news text \\ using formal syntactic theory}

\author{
    Olga Zamaraeva\textsuperscript{1} \\ \texttt{olga.zamaraeva} \And
    \hspace{-3em} Dan Flickinger\textsuperscript{2} \\ \hspace{-3em}\texttt{danflick} \And
    \hspace{-4em}Francis Bond\textsuperscript{3} \\ \hspace{-4.5em}\texttt{francis.bond} \And
    \hspace{-3em}Carlos Gómez-Rodríguez\textsuperscript{1} \\ \hspace{-3em}\texttt{carlos.gomez} \\ \AND\\[-1cm]
    \textsuperscript{1}Universidade da Coruña, CITIC (\texttt{@udc.es})\\
    \textsuperscript{2}Independent Researcher (\texttt{@alumni.stanford.edu}) \\
    \textsuperscript{3}Palacký University at Olomouc, Department of Asian Studies (\texttt{@upol.cz}) \\
  }

\begin{document}
\maketitle
\begin{abstract}
This study provides the first comprehensive comparison of New York Times-style text generated by six large language models against real, human-authored NYT writing. The comparison is based on a formal syntactic theory. We use Head-driven Phrase Structure Grammar (HPSG) to analyze the grammatical structure of the texts. We then investigate and illustrate the differences in the distributions of HPSG grammar types, revealing systematic distinctions between human and LLM-generated writing. These findings contribute to a deeper understanding of the syntactic behavior of LLMs as well as humans, within the NYT genre.
  
\end{abstract}

\section{Introduction}
Studying linguistic properties of LLM-generated text and comparing it to human-authored text is a topic of growing interest in the field of natural language processing (NLP). Previous research has predominantly focused on training classifiers (is the text LLM-generated or no); a few studies include an analysis of differences in vocabulary distribution, use of dependency structures, or sentiment properties of the text (see \S~\ref{sec:related}). In this study, we systematically analyze \emph{grammatical} differences of LLM-generated vs.\ human-authored text through the lens of a formal syntactic theory developed for linguistic research independently of NLP.\footnote{Annotation schemes such as Universal Dependencies \citep[UD:][]{nivre2016universal} or Penn Treebank \cite[PTB:][]{ptb:1993} are related to syntactic theory but they have been developed as guidelines for hand-annotating corpora specifically for NLP. As such, they are less detailed and consistent than a formal theory and less independent from NLP tasks themselves.} Using a formal theory for analysis and evaluation is a way to overcome some of the biases that arise from using tools developed directly in the context of designing NLP tasks.  We hope this will lead to further systematic discoveries about grammatical properties of LLM-generated text and how they differ from human-authored text. In this paper, we use the broad-coverage English Resource Grammar \citep{flickinger2000building, Flickinger:11} to analyze texts in the New York Times genre.


\section{Related work}
\label{sec:related}

Our study is concerned with the analysis of the grammatical properties of LLM-generated texts as compared to human-authored texts. Here, we review the literature with a similar focus. This leaves out of scope papers concerned with building classifiers or with sentiment and semantic analysis. 

\citealt{munoz2024contrasting}  include a study of syntactic and vocabulary diversity in NYT-style news. They conclude that measurable differences can be detected, including at the level of grammar, and that human-authored texts exhibit more variety of vocabulary, shorter constituents, and more optimized dependency distances.   \citet{narayanan2024explaining} use the Universal Sentence Encoder \citep[USE:][]{cer2018universal} to compare human-authored and AI-generated code explanations and find statistical differences, though without linguistic analysis. \citet{sandler2024linguistic} base the comparison on ChatGPT-human dialogues, using primarily lexical features, not syntactic, and find greater diversity in texts written by humans. Notably, they use dictionary-style features and not just raw vocabulary. So do \citet{alvero2024large}, who compare college application essays (submitted in 2016-2017) with texts generated by GPT-3.5 and GPT-4. They find that human authors show more variety in e.g.\ verb usage. \citet{juzek2025does} study the vocabulary of LLMs linking it to the increase of use in certain vocabulary items in scientific abstracts (e.g.\ the word `delve').   \citet{park2025does} perform a statistical comparison by clustering linguistic features (this is necessary to obtain statistically significant results in the context of multiple comparisons between many features). 
They conclude that LLM-generated texts have a distinct statistical footprint from human-authored text. \citet{shaib2024detection} compare strings of POS-tags, which they call ``syntactic templates'', finding that LLMs tend to repeat these templates more than humans do. Finally, several studies base the comparison on a set of linguistic features proposed for rhetoric styles by \citeauthor{biber1991variation} (\citeyear{biber1991variation}, \citeyear{biber1995dimensions}) and \citet{biber2019register}. In particular, \citet{reinhart2024llms} show that LLMs prefer certain grammatical constructions and thus struggle to match styles that do not employ them (according to \citeauthor{biber1991variation}). The constructions include participial clauses, `that'-subject clauses, nominalization, phrasal and clause coordination. \citealt{sardinha2024ai} also uses the ``Biber features''. This study is perhaps the closest in spirit to ours, since it uses an independently developed linguistic framework and presents examples of the differences found. 





\section{Methodology} 

\label{sec:background}
The central idea of our methodology is to apply formal syntactic theory to analyzing structural properties of texts generated by LLMs as compared to human-authored texts. We use the HPSG theory of syntax (\S\ref{sec:hpsg}), specifically its implementation as the English Resource Grammar (\S\ref{sec:erg}), the largest available implementation of a formal grammar in terms of its coverage over naturally occurring text (in any language and in any theory). While applying such methodology implies the investment in building the grammar, the HPSG theory and the formalism were developed precisely to be used for a wide variety of languages. The cross-linguistic applicability of the theory has been continuously tested in the context of the Grammar Matrix \citep{Ben:Fli:Oep:02, Ben:Dre:Fok:Pou:Sal:10, zamaraeva202220} and the AGGREGATION \cite{bender:agg:2020, howell2022building} projects. 

\subsection{HPSG}
\label{sec:hpsg}
Head-driven Phrase Structure Grammar \citep[HPSG:][]{Pol:Sag:94} is a formal theory of syntax that uses a fully explicit formalism, so it can be implemented on the computer in its entirety as a grammar which then maps sentences to complete structures automatically, while remaining fully consistent and interpretable. The theory represents syntactic structure and elements of the syntax-semantic interface (dependencies, quantifier scope, information structure) as a complex graph, which can also be visualized as an attribute-value matrix of features and their values (such as the feature {\sc head} having a value \emph{noun}). HPSG assumes lexical types which can house multiple lexical entries, and, unlike raw vocabulary forms, lexical types contain information about syntactic properties of words. 

\begin{figure}[h!]
    \centering
    \small
    \begin{forest}
    [\emph{noun}\\\begin{avm}
\[LOCAL\|HEAD \emph{noun} \]
\end{avm} [\emph{clausal complement}\\\begin{avm}
\[COMPS \emph{nonempty} \]
\end{avm}[law-n2]][\emph{mass-count}\\\begin{avm}
\[COMPS \emph{empty} \]
\end{avm}[law-n1]]]
    \end{forest}
    \caption{Part of the HPSG type hierarchy (simplified; adapted from ERG).}
    \label{fig:hier}
\end{figure}
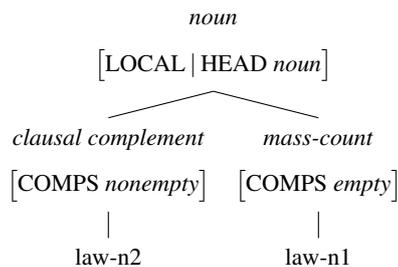

The grammar as a whole (the lexicon included) is a hierarchy of types. Figure \ref{fig:hier} shows a very small and simplified portion of the HPSG type hierarchy, with only two features ({\sc head} and {\sc comps}, complement list). This part pertains to the lexicon and lexical types. The noun `law' can behave in different ways syntactically, which motivates two lexical entries belonging to two different types (which may house other nouns as well). In \S\ref{sec:results} we report on how this word is one of the examples of differences in human-authored and LLM-generated texts that we examined. The real type hierarchy, such as the one in the ERG (\S\ref{sec:erg}), consists of hundreds of types with dozens of features, allowing us to examine grammatical properties of sentences in detail. 

\begin{table*}[ht!]
    \centering
        \caption{Datasets: reproduced in full from Table 1 in \citealt{munoz2024contrasting}, plus the information on Redwoods.}

\begin{adjustbox}{max width=\textwidth}
    \begin{tabular}{lrrrl}
        \toprule
        Dataset & \# Sent. in dataset & Model size & Training tokens & Data sources \\
        \midrule
         {\multirow{4}{*}{{LLaMa}}} & 37,825 & 7B & 1T & English CommonCrawl (67\%), C4 (15\%), \\
          & 37,800 & 13B & 1T & GitHub (4.5\%), Wikipedia (4.5\%), \\
         & 37,568 & 30B & 1.5T & Gutenberg and Books3 (4.5\%), ArXiv (2.5\%),\\
         & 38,107& 65B & 1.5T & Stack Exchange (2\%)\\
        \midrule
        {\multirow{3}{*}{{Falcon}}} & & & & RefinedWeb-English (76\%), RefinedWeb-Euro (8\%), \\
        & 27,769 & 7B & 1.5T & Gutenberg (6\%), Conversations (5\%) \\
        & & & & GitHub (3\%), Technical (2\%) \\
        \midrule
        Mistral & 35,086 & 7B &  Not disclosed & Not disclosed \\
        \midrule
        Original NYT & 26,102 & N/A & N/A & New York Times Archive, Oct.\ 1, 2023 - Jan.\ 24, 2024 \\ 
        \midrule
        Redwoods (WSJ) & 43,043 & N/A & N/A & Wall Street Journal sections 1-21 \\ \cline{2-5}
        Redwoods (Wikipedia) & 10,726 & N/A & N/A & Wikipedia \\
        \bottomrule
    \end{tabular}
    \label{tab:dataset-sizes}
    \end{adjustbox}
\end{table*}

\subsection{English Resource Grammar}
\label{sec:erg}
The English Resource grammar is a grammar of English implemented in HPSG \citep{flickinger2000building, Flickinger:11}.\footnote{Regular releases: \url{https://github.com/delph-in/erg}} The ERG is continuously developed as part of the DELPH-IN open-source grammar engineering initiative.\footnote{\url{https://github.com/delph-in/docs/wiki}}  It is a broad coverage precision grammar, meaning that it will parse 94\%\footnote{Per the 2025 release documentation} of reasonably well-edited English text but is not expected to yield any structure for a sentence impossible in English. Since the grammar is precise and consistent, it can be used to automatically create precise and consistent treebanks. It has been shown that including such treebanks in the training data improves performance of various NLP systems \citep{lin2022towards, hajdik2019neural, chen-etal-2018-accurate, buys2017robust}. Some of the properties of the ERG are summarized in \S\ref{sec:data}, Table~\ref{tab:erg}. The grammar is implemented in the DELPH-IN Joint Reference Formalism \citep{Copestake:02:CLE} and can be used with any DELPH-IN tools. We parsed the data with the latest version of the ERG\footnote{\url{https://github.com/delph-in/erg/releases/tag/2025}} and ACE \citep{crysmann2012towards},\footnote{\url{https://sweaglesw.org/linguistics/ace/download/ace-0.9.34-x86-64.tar.gz}} and then used the Pydelphin tools\footnote{\url{https://pydelphin.readthedocs.io/}} along with packages such as Numpy \citep{harris2020array}, Pandas \citep{mckinney-proc-scipy-2010}, and scikit-learn \citep{scikit-learn} to analyze the derivations by counting the occurrences of phrasal constructions, lexical (inflectional and derivational) rules, and lexical types, and studying the relative frequency distributions through cosine similarity and diversity metrics (see \S\ref{sec:results}). 


\section{Data and generative models}
\label{sec:data}
To study the differences between LLM-generated and human-authored news texts, we use the dataset created by \citet{munoz2024contrasting}. We choose this dataset for two main reasons: 1) by using news articles, we can make sure the LLMs did not have access to the corresponding human-authored articles at the time of training; 2) by reusing the dataset from a previous study, we enable comparisons of analyzing the data with UD and with the fully-fledged grammatical theory provided by HPSG. 
In addition, we used the Wall Street Journal (WSJ) and Wikipedia portions of the Redwoods Treebank \citep{oepen2004lingo}, an ERG-parsed corpus accompanying each release of the ERG. We use WSJ and Wikipedia to see which differences between human and LLM writing persist beyond the NYT style. We release the ERG-parsed LLM-generated data through GitHub.\footnote{\url{https://github.com/olzama/llm-syntax/releases/tag/1.0.0}} 

\begin{figure*}[t]
\centering
\begin{minipage}[t]{0.48\textwidth}
    \centering
    \includegraphics[width=\linewidth]{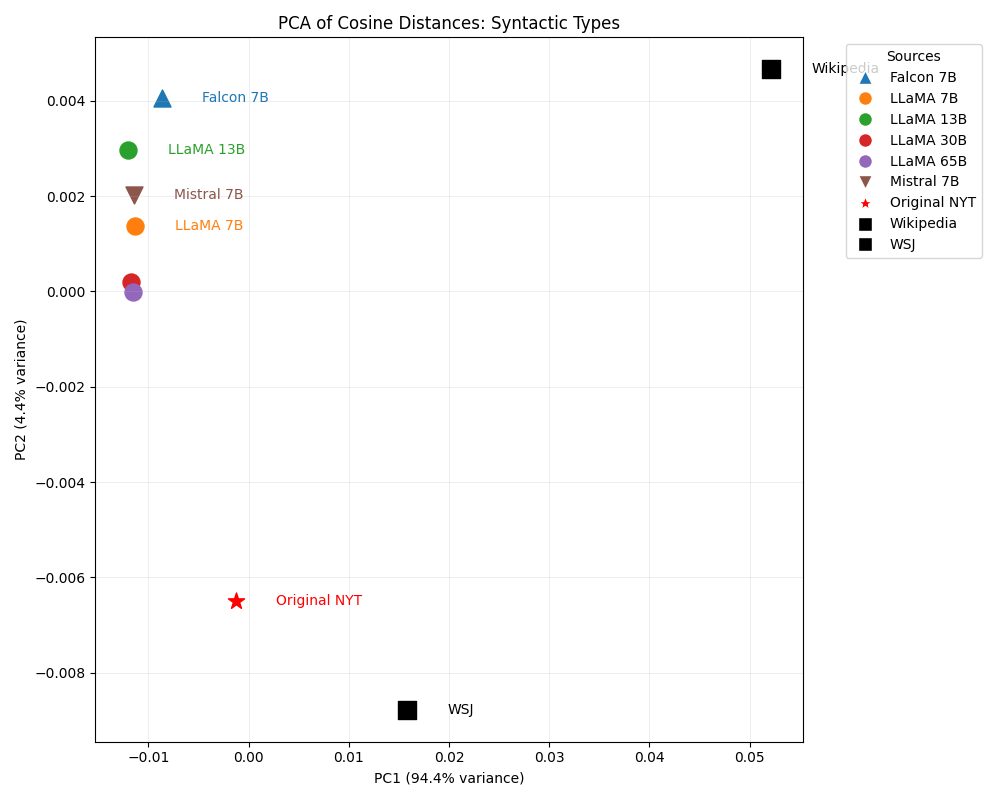}
    \caption{Cosine similarity: syntactic types}
    \label{fig:pca1}
\end{minipage}
\hfill
\begin{minipage}[t]{0.48\textwidth}
    \centering
    \includegraphics[width=\linewidth]{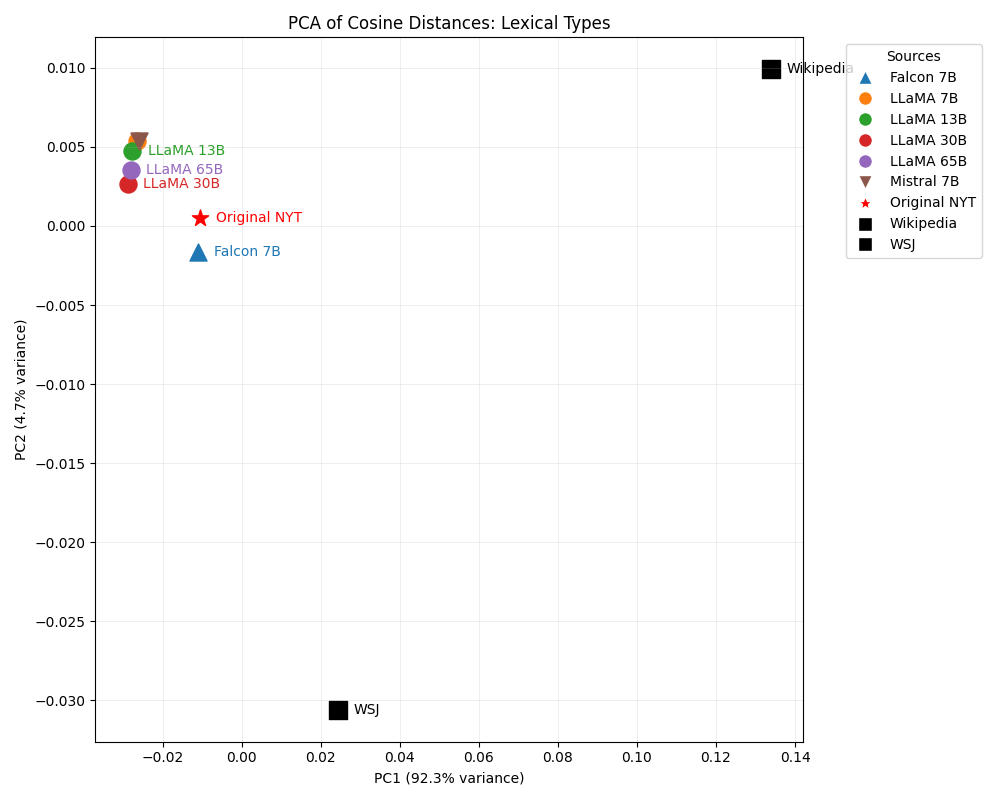}
    \caption{Cosine similarity: lexical types}
    \label{fig:pca2}
\end{minipage}
\end{figure*}

\begin{figure}[h!]
    \centering
    \includegraphics[width=\linewidth]{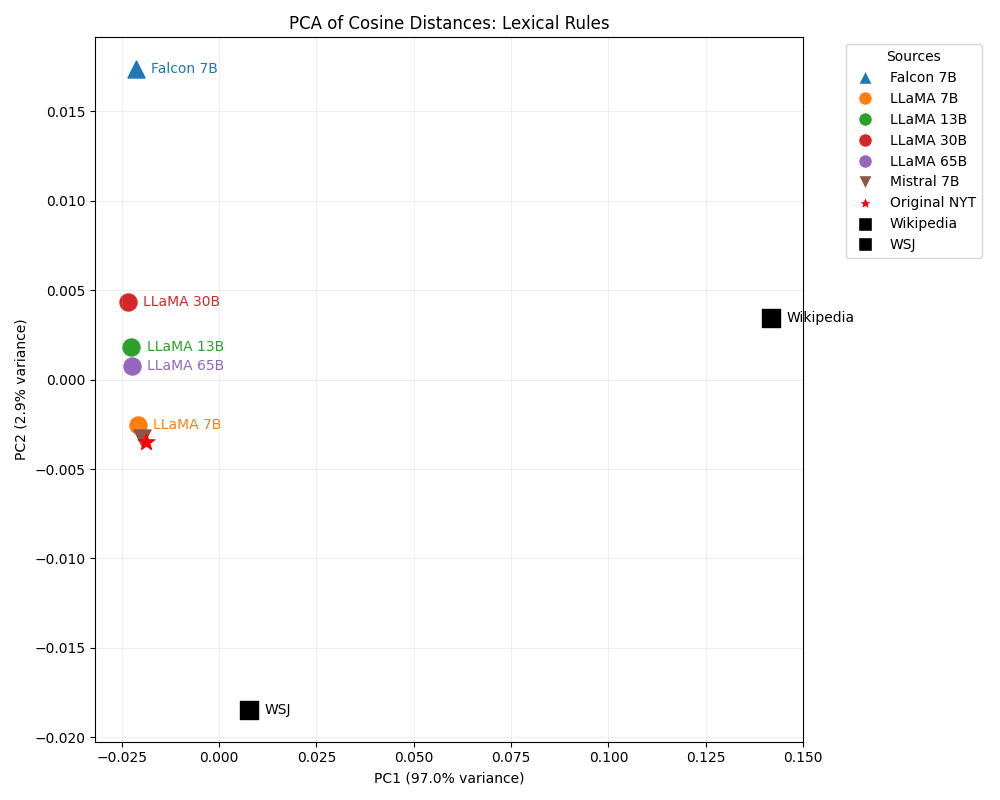}
    \caption{Cosine similarity: lexical rules}
    \label{fig:pca3}
\end{figure}

The `NYT' datasets from \citealt{munoz2024contrasting} include the original New York Times (NYT) article lead paragraphs and LLM-generated texts obtained from 6 different LLMs by prompting them with the headlines together with the first 3 words of the lead paragraph.\footnote{The LLM-generated data associated with \citealt{munoz2024contrasting} can be found here: \url{https:// zenodo.org/records/11186264}} The original NYT human-authored data consists of the lead paragraphs for articles between October 1, 2023, and January 24, 2024 obtained with the NYT Archive API.\footnote{\url{https://developer.nytimes.com/docs/archive-product/1/overview}} The LLMs they used were all released prior to October 1, 2023, and included various versions of LLaMA \citep{touvron2023llama}, the 7B version of Falcon \citep{almazrouei2023falcon}, and the 7B version of Mistral \citep{jiang2023mistral}. Following \citet{munoz2024contrasting}, we want to consider the influence of scaling (different LLaMas with the same architecture, training dataset and training setup, but different model size) separately from the other aspects that differentiate the LLMs (LLaMa vs Mistral vs Falcon). The properties of the datasets and the models used to generate them (where appropriate) are in Table~\ref{tab:dataset-sizes}.  LLM-generated datasets have more sentences, but the sentences written by humans are longer (see Figure 3 in \citealt{munoz2024contrasting}).

\begin{table}[tbp]
\centering
\small
    \begin{tabular}{lrr}
    Construction type & ERG & Data\\
    \hline
        syntactic & 298 & 289\\
        lexical type & 1,398 & 1,105\\
        lexical entry & 44,366 &27,311\\
        morphological rule & 100 & 99 \\
    \end{tabular}
    \caption{Properties of the English Resource Grammar and the coverage of types by the NYT data}
    \label{tab:erg}
\end{table}
The NYT data accounts for almost all syntactic and morphological rules registered in the grammar; for about 79\% of the lexical types, and for about 61\% of the lexical entries (Table~\ref{tab:erg}).

\section{Results}
\label{sec:results}

We present the comparison of type distributions between the human-authored and LLM-generated data, including WSJ and Wikipedia data to see whether the differences persist across styles or genres. We look at cosine similarity of the construction distributions (\S\ref{sec:res:cosines}) and at two diversity indices (\S\ref{sec:diversity}). We look at syntactic and lexical types as well as lexical (morphological) rules separately.

\begin{figure*}[h!]
    \centering
    \includegraphics[width=\textwidth]{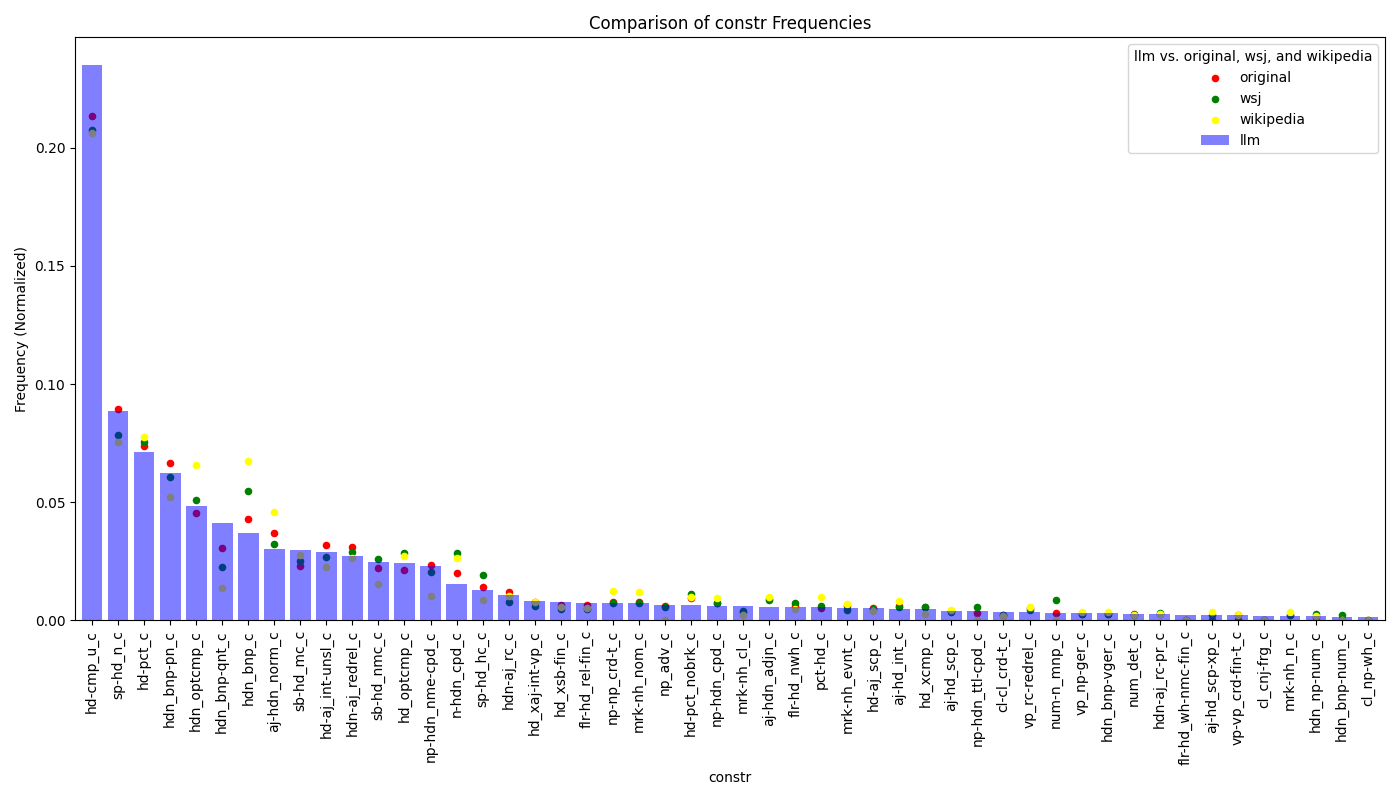}
    \caption{LLM use of syntactic constructions compared to human writers. Cases of particular interest are where the dots cluster closely together and are noticeably higher or lower than the blue bar representing LLM. Also of certain interest are cases where all the dots are higher or lower than the bar but not very close to each other.}
    \label{fig:frequent-constr}
\end{figure*}

\subsection{Cosine similarity}
\label{sec:res:cosines}
We find that human authors and LLMs clearly differ in terms of their corresponding syntactic and lexical type distributions, and that this may persist across style and genre.\footnote{We use PCA projection to help visualize the differences in the 98-100\% similarity range. The underlying data is provided in Appendix \ref{appendix:b}.} If we consider only syntactic and lexical types (Figures \ref{fig:pca1}-\ref{fig:pca2}),\footnote{The data is not directly comparable, hence the scale differences.} we see clearly that human-authored texts are distinct in their HPSG type distributions from the closely-clustered LLMs and furthermore, that human-authored NYT texts are more similar to WSJ (different style, same genre) than to Wikipedia (different genre). This is true for syntactic and lexical types, although with lexical types, Falcon is an outlier, and the effect of style and genre seems bigger. However, in terms of lexical (inflectional and derivational) rules, we observe that the distribution of human NYT authors is very similar to LLMs except Falcon. These findings align with what we see when we apply diversity metrics (\S\ref{sec:diversity}). In this paper, we focus on the most salient differences between LLMs and human NYT authors, and investigating the intriguing role of lexical rules remains future work. One hypothesis is that the distribution of lexical rules is very closely tied to genre and style (and that the Falcon model is somehow special in this respect).





\subsubsection{Frequent syntactic constructions}
\label{sec:res:syntax}

Among the frequent syntactic constructions (Figure~\ref{fig:frequent-constr}; Appendix \ref{sec:appendix:erg}), we see differences insensitive to genre\footnote{We have run the Mann-Whitney U-test for statistical significance for these comparisons. The p-values < 0.05 are listed in Appendix \ref{sec:appendix-d}. However, we perform a large number of comparisons, and when we apply FDR correction to the p-values, none of them come out as significant, which is not surprising given that we only have 9 datasets to compare.} in the head-complement construction (human authors use less of it in all human-authored datasets we examined), a couple of punctuation-related constructions (note that from the point of view of the ERG, punctuation is not only a token; it also matters how exactly it gets placed in the sentence, so, this is a syntactic matter), and the adjunct-head construction licensing double modification (e.g.\ \emph{big old cat}). There might be something of note going on with bare noun phrases and noun compounds as well; the human authors appear to use them more; however the differences between styles (WSJ) and genre (Wikipedia) seem to be greater than the differences between human NYT writers and LLM-generated NYT-style news. Differences in punctuation have been observed \citep{munoz2024contrasting}; however the head-complement construction is a general grammatical feature which does not have a direct equivalent in the UD framework. In UD, there is the {\sc obj} dependency, which refers to a dependency between a direct object and a verb, and is a concept from the syntax-semantic interface. A head-complement construction is a general syntactic construction that licenses constituents which combine a head element with its complement. The head does not need to be a verb (nouns and adjectives can have complements too, for example). In this study, we do not include further analysis of the differences in the use of head-complement constructions by LLMs and by human authors, but in future work, it would be interesting to see, for example, whether there is a difference in subconstituents or in the lexical types or entries forming the head-complement constituent itself. 

\begin{figure*}[t]
\centering
\begin{minipage}[t]{0.48\textwidth}
    \centering
    \includegraphics[width=\linewidth]{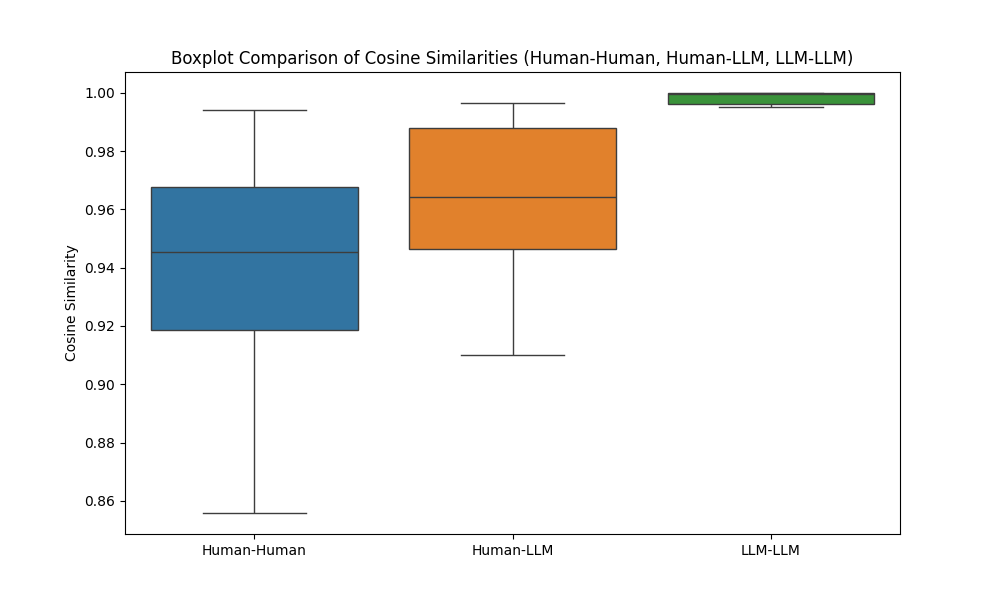}
    \caption{Humans vary more from one another than they do from an LLM, and LLMs vary little from each other.}
    \label{fig:pair-compar-all}
\end{minipage}
\hfill
\begin{minipage}[t]{0.48\textwidth}
    \centering
    \includegraphics[width=\linewidth]{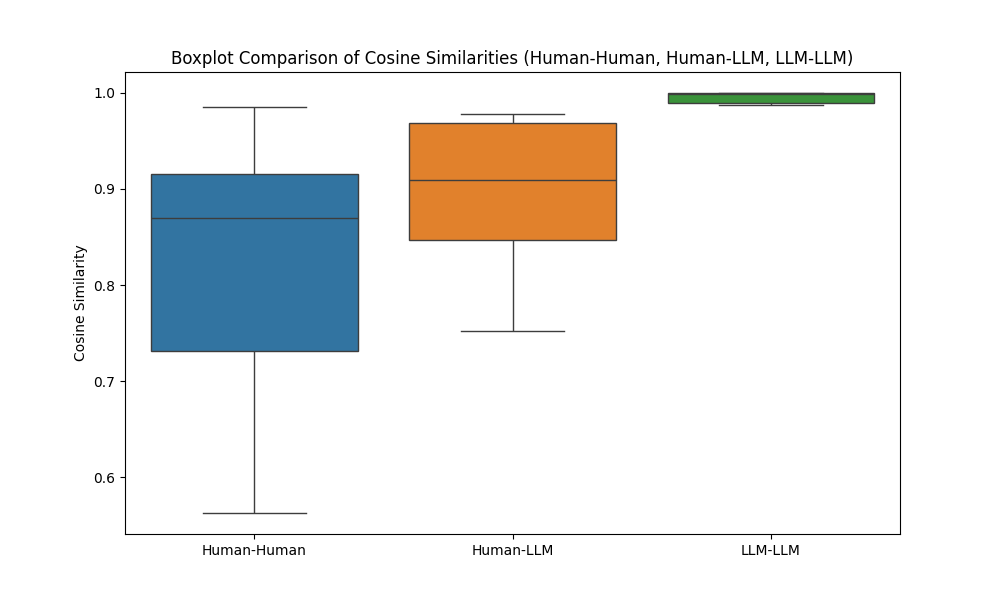}
    \caption{Human authors have particularly large variance when it comes to the lexical types they use}
    \label{fig:pair-compar-lextype}
\end{minipage}
\end{figure*}

\subsubsection{Syntactic long tail}
It is possible that some salient differences lie in the ``long tail'' of the distributions (not shown in Figure \ref{fig:frequent-constr}). The ERG is a unique resource to study this long tail, being a comprehensive representation of the English language which, while validated empirically, was developed with close attention to a wide range of phenomena, not only the most frequently occurring ones.

The following constructions occur only 0 or 1 times in a sample from human-authored NYT text, while similar size samples from the LLM-generated texts contain more than 10 instances: sequence of numbers; fragment lexical conjunction (``But!''); parenthetical modifier (``Some person (tall) was running away''); mass noun coordination (`sand and gravel');\footnote{Note the special syntactic properties of this construction, such as underspecified number agreement: \emph{Sand and gravel has/have arrived.}} modifier phrase formed from `measure' nouns (\emph{We slept the last mile}). 

Humans use all of these long-tail constructions occasionally (which is how they came to be represented in the ERG in the first place); their not occurring in the NYT dataset could just be by chance.  Future experiments with more data are needed. In the meantime, we show that HPSG analysis aligns with previous findings with UD (e.g.\ that current 
 LLMs are known to favor numbers and measure-related vocabulary \citep{munoz2024contrasting}), and identify constructions possibly typical for LLMs which have not previously been noted (\S\ref{sec:linganalysis-syntax}).

\subsubsection{Lexical (morphological) rules}
We do not observe any differences of note in the LLM and human use of frequent lexical rules (inflectional and derivational morphology),\footnote{The only infrequent rule of note is the one related to currencies (``A one-dollar book'', where the rule is responsible for the special currency-related properties of the phrase ``one dollar'', as compared to any generic noun phrase).} except in all human-authored datasets, plural nouns have been used with greater relative frequency than in the LLM-generated texts (but there is more variation between the genre/style). This shows once again the importance of separating morphological information from syntactic and lexical when analyzing language (cf.\ \citealt{bender2005implementation}). 

\begin{figure*}[t]
\centering
\begin{minipage}[t]{0.48\textwidth}
    \centering
    \includegraphics[width=\linewidth]{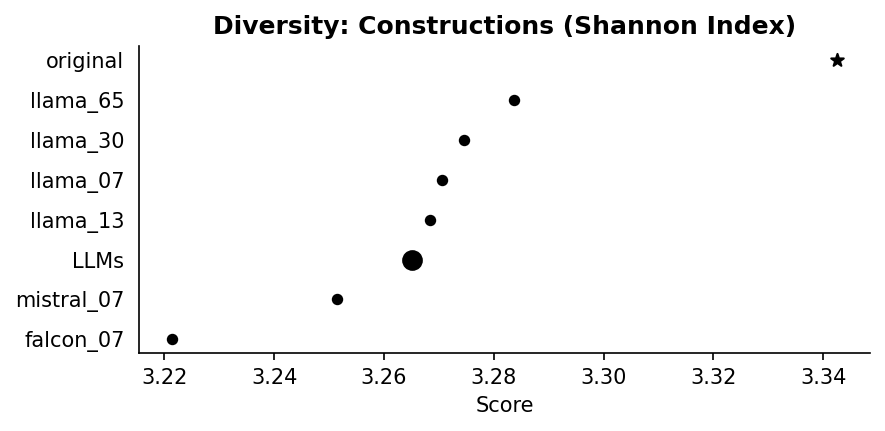}
    \caption{Construction Diversity (Shannon Index)}
    \label{fig:shannon-cc}
\end{minipage}
\hfill
\begin{minipage}[t]{0.48\textwidth}
    \centering
    \includegraphics[width=\linewidth]{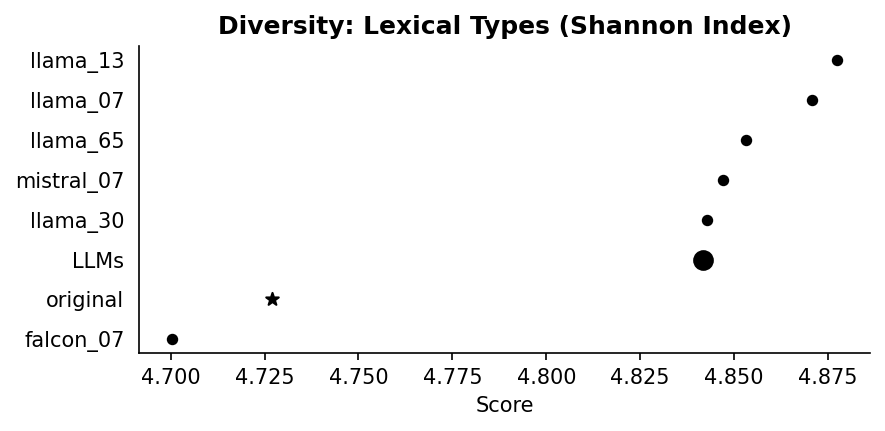}
    \caption{Lexical Type Diversity (Shannon Index)}
    \label{fig:shannon-lt}
\end{minipage}
\end{figure*}

\subsubsection{Lexical entries and types}
\label{sec:res:lexical}

Human writers use roughly twice as many different lexical \emph{entries} as each LLM taken separately (Table~\ref{tab:combined_lex}). This confirms previous findings that humans show more variation in vocabulary use (see \S\ref{sec:related}). But if we combine all of the LLM-generated data and sample from it, this collective LLM author has a greater lexical diversity than the human authors. This calls for further investigation of what makes the collective LLM vocabulary more varied. As for lexical \emph{types}, LLMs seem to have greater diversity in terms of just the number of unique lexical types they use in the sample (with the exception of falcon-7B). When we look at the specific lexical types accounting for these distinct footprints, we see that of the 66 types which do not occur in any of the LLMs, 43 belong to the bottom 10\% in terms of frequency, 21 to the bottom 25\%, and only 2 to the bottom 50\%. The two frequent ones include a special kind of mass noun such as `next' in `The next is Kim', and the special kind of `if' such as in `The happy if confused customer left' (the customer was confused, but was happy nevertheless).  

\begin{table}[tbp]
    \centering
    \small
    \begin{tabular}{lrrrr}
        \textbf{Model} & \multicolumn{2}{c}{\textbf{Lexical Types}} & \multicolumn{2}{c}{\textbf{Lexical Entries}} \\
         & Not in & Only in & Not in & Only in  \\
        \hline
        llama 7B & 62 & 70 & 5,704 & 2,519 \\
        llama 13B & 71 & 80 & 5,557 & 2,617 \\ 
        llama 30B & 65 & 62 & 5,531 & 2,608 \\ 
        llama 65B & 66 & 74 & 5,302 & 2,745 \\
        mistral 7B & 73 & 76 & 5,809 & 2,353 \\
        falcon 7B & 91 & 55 & 6,212 & 2,015 \\
        all llms & 66 & 70 & 1,721 & 2,398 \\
    \end{tabular}
    \caption{Lexical types and entries found only in human-authored or only in synthetic data, sample 25K.}
    \label{tab:combined_lex}
\end{table}

\subsubsection{Individual author variance}
\label{sec:pairwise}
In addition to looking at NYT human authors collectively, we are interested in how much they differ from each other and whether these individual differences are greater or not than the differences between humans and LLMs (Figures \ref{fig:pair-compar-all}-\ref{fig:pair-compar-lextype}). We perform the comparison with 12 authors that have more than 100 sentences attributed to them in the NYT data. The comparison is again based on cosine similarity, where the vectors are construction/type frequencies normalized by total number of occurrences in the data. Here we include a comparison based on all the HPSG types together.

We find that human writers differ from each other more than a human author differs from an LLM, and LLMs differ very little from each other (Figure \ref{fig:pair-compar-all}). If we look at lexical types, we see that humans vary particularly strongly in their use of lexical types, while LLMs have the same kind of small variance in this respect as they do in other types of constructions (Figure~\ref{fig:pair-compar-lextype}).\footnote{Since we have more data for each LLM than for each human, we confirmed that we see similar distributions in a balanced dataset, if we sample randomly from the LLM data.} 

As far as we know, our study is the first pairwise comparison of human authors and LLMs along detailed grammatical dimensions, and we show for the first time that a human-authored text is more similar to an LLM-generated text than to another human-authored text (by a different author). This makes sense if we see an LLM-generated text as ``averaged'' with respect to grammatical features that humans use in their language. This can also be seen in their increased use of the most general structures such as the head-complement phrase (Figure \ref{fig:frequent-constr}). 
Our results also confirm the previous observations that LLMs are very similar to each other in terms of the types of constructions that they use (see \S\ref{sec:related}).

\subsection{Diversity}
\label{sec:diversity}

To quantify diversity in the texts we applied two biodiversity
measures that have become standard in stylometry and authorship
attribution \citep{mckinney-proc-scipy-2010,stamatatos2009survey}:
\textbf{Shannon entropy} $H$ and the \textbf{Gini--Simpson index}
$1-\lambda$.  The former captures the balance (evenness) of the
distribution, while the latter is interpretable as the probability
that two randomly drawn tokens belong to different types.  Because
both indices give the same rank orderings in our data (see
Appendix~\ref{appendix:c}), we only discuss Shannon entropy here.

\paragraph{Constructions}
Figure~\ref{fig:shannon-cc} plots $H$ for syntactic constructions.
Human‐produced texts (``original'') are clearly the most diverse
($H=3.342$), and all language-model outputs fall below that benchmark
($H=3.221$–$3.284$).\footnote{A permutation test with 10,000 resamples confirms a reliable gap ($p<0.01$).}
The \textit{largest} LLaMa model (65 B) is the closest to humans
($H=3.284$), whereas the Falcon model is the least diverse
($H=3.221$).  Interestingly, when we pool every LLM output into a single corpus, its diversity \emph{drops} slightly to $H=3.265$.  Aggregation
adds a handful of rare constructions that were unique to individual models,
but it also amplifies the high-frequency, general constructions that all
models share, skewing the distribution and lowering overall entropy.

\paragraph{Lexical types}
The pattern reverses when we consider lexical types (Figure~\ref{fig:shannon-lt}).  
Here, LLM outputs are more diverse than human-authored texts: the least diverse system (Falcon) scores $H = 4.700$, followed by the original human data at $H = 4.727$. All other LLMs surpass humans, with LLaMA-13B at the top ($H = 4.877$). These differences are statistically significant ($p < 0.01$).   Investigating this pattern reversal is future work.


 %
 


\begin{table*}[h!]
    \begin{adjustbox}{width=\textwidth}
    \centering
    \begin{tabular}{llrr}
       Construction  &  Ex & Humans & LLMs (avg) \\
       \hline
       Absolute VP  & `As told, ...' & 10 & 3.8 \\
       Double NP apposition & `an eye for detail, decades of a culture in transition' & 11 & 5.2\\
        Double appos.\ modifier& `accurate, but inadequate, descriptor' &12 &5.6\\
       Adjective-participle modifier & `right-handed', `red-colored' &125 & 64.6\\
       Bare NP coordination & `..., author and commentator, ...' & 311 & 117 \\
       Paired marker & `Both this article and other discussions', `not only...' & 326& 185\\ 
       Adjective coordination & `emotional and spiritual' & 390 & 625\\
       Modifier clause appos.\ &`his critics, mostly unnamed' & 826 &434\\
       Participial clause &`...having tried that,...' & 1,736& 1,116\\ 
       \hline 
       Inverted adjunct & `Below are some of the facts...' &5& 14.8 \\
       Clause-clause coordination & `which ones are and which ones aren't' & 45 & 105 \\
       
       Filler-head non-question wh & `How best to proceed: [...]' & 149 & 306 \\ 
       Questions & `How do you stay safe?' & 268 & 428 \\
       Clause conjunction fragment & `But the observation suits him.' &939 &2,076 \\
       Marker clause & `..., and that's a good thing' & 2,891& 5,660\\
       Relative clauses & `...a vote that many in Europe have seen as a bellwether or support...' & 4,929& 6,721  \\
       Clause with extracted subject & `Chris Snow, [...], became an advocate for the victims of the disease.'& 5,072 & 7,327 \\
       Subject-head & `The house passed the measure earlier this week.'& 17,850 & 27,753\\
       Quantity NP & `many in Europe' & 23,611&40,881 \\

       Head-complement & `It's not acceptable for democracy'& 164,806 &224,529 \\
    \end{tabular}
        \end{adjustbox}

    \caption{Examples of selected syntactic constructions which seem to have noticeably different frequency in human-authored and in LLM-generated data (25K sentence sample)}
    \label{tab:example-syntax}
\end{table*}

\section{Examples of salient differences}
\label{sec:ling-analysis}
\subsection{Syntactic constructions}
\label{sec:linganalysis-syntax}
We have examined some of the constructions which are used noticeably more by human authors than by the collective LLM, or vice versa.\footnote{We have selected such constructions based on the statistical significance of the comparison between relative frequencies.} The constructions where the difference in relative frequency is most clear notably include the head-complement construction and the subject-head construction\,---\,the two most basic constructions forming any typical clause. Here we do not attempt to analyze the numerous examples of this kind of construction use (leaving it to future work) but nonetheless include an example from the corpus for each (Table~\ref{tab:example-syntax}).

Table~\ref{tab:example-syntax} aligns with some of the previous findings (\citealt{munoz2024contrasting} and \citealt{sardinha2024ai}, among others), namely that LLMs tend to use more quantity-related words and phrases; that LLM-generated texts have more structures which can be classified as a generic `verb phrase' (VP) or `sentence' (S), which in our analysis would correlate with the higher frequencies of head-complement and head-subject constructions; that LLMs tend to use more clause coordination; and that human authors tend to produce more prepositional phrases in the NYT-style writing. However, we do not confirm the finding of \citet{sardinha2024ai} that LLMs use more participial modifiers; in our data, humans use it more. In addition, we can hypothesize several other systematic differences using the ERG elaborate syntactic type hierarchy. According to our analysis, the LLMs collectively tend to use more relative clauses and questions, more clause chains, more clauses with extraposed subjects, and more extraposed adjuncts. In contrast, human authors use more stylistic devices such as participial modifiers, full clause modifiers, double adjective apposition, coordinated prepositional phrases, coordinated adjective modifiers, double noun phrase apposition, and the so-called absolute verb phrase. In summary, human authors use more of the lower frequency, stylistically special constructions. 

\subsection{Lexical types and lexical entries}
\label{sec:linganalysis-lextypes}

There are many differences between the lexical footprints of LLM-generated and human-authored text in terms of low-frequency items. If the word is both low frequency and belongs to a lexical type which does not have many members, it is hard to say whether its use is just an accident or could be informative. Therefore, we focus on items which are high frequency but occur only in human-authored or only in LLM-generated data (Tables \ref{tab:human-lexentry-unique}-\ref{tab:llm-lexentry-unique}).\footnote{We must note that such differences can always be attributed to sampling. Obviously, a human writer can use any of the items from Table~\ref{tab:llm-lexentry-unique}, and it is trivial to have an LLM produce any of the things from Table~\ref{tab:human-lexentry-unique}.} 

We take advantage of the ERG lexical type hierarchy and look at how the lexical entries which seem to distinguish LLM-generated text from human-authored text can be grouped together in grammatical terms.
 One example of the lexical entries found only in human-authored data is `law\_n2' (with a clausal complement). This lexical entry is present in the ERG lexicon along with the mass-count noun `law\_n1' and belongs to a different lexical type. The word `law' certainly occurs in LLM-generated data as well, but only as the mass-count noun. We find that only in the human-authored data is this word used as something that can take a clausal complement, e.g.\ `There is a law that...'. This is the kind of distinction that we are looking for in our study; if we did not have the ERG lexicon at our disposal, we could overlook the distinction.\footnote{Of course such distinctions should correlate with the differences in syntactic construction use.}

\begin{table}[t!]
    \centering
    \small
    \begin{tabular}{lrl}
       Lex.\ entry & occurr.\ & example \\
       \hline
        OOV verb & 178 & `twerk', `steamroll'\\
       risk\_n3 & 144 & `at your own risk'\\
        haven't & 88 & `If you haven't already...' \\ 
        night\_def & 82 & `spend the night'\\
        a\_per\_p & 81 & `a night', `a barrel'\\
        see\_imp & 69 & `See the results...'\\
        including\_pp & 65 & `...including on April 17'\\
        yet\_conj & 64 & `...yet there it is' \\
        dozen\_a1 & 62 & `a couple dozen pages' \\
        winter\_n1 & 61 & `With winter approaching,..'\\
        down\_vmod & 59 & `walk/skip/sprint down'\\
        almost\_deg2 & 56 & `almost always' \\
        present\_v1 & 51 & `A puzzle presented to students'\\
        over\_pp & 50 & `The wait is over.'\\
        black\_n2 & 50 & `growing up Black'\\
        
    \end{tabular}
    \caption{Frequent (top 15) lexical entry usages unique for the human-authored dataset}
    \label{tab:human-lexentry-unique}
\end{table}

One of the main things that we see in Table~\ref{tab:human-lexentry-unique} is that the real (human) authors of the NYT use more informal language even though they are following a style guide. A LLM certainly could also use expressions like `a couple dozen' and `haven't', and in fact it does use `won't' and `that's' (Table~\ref{tab:llm-lexentry-unique}), but overall each LLM seems to be more consistently adhering to the style of the prompt. Another trait of the human-authored data is more direct/strong language, such as imperatives and expressions such as `at your own risk'.\footnote{We have no ready explanation for why the word `winter' (without the article) would only occur in human-authored data, or why the verb `realize', in its most common usage, would happen to not occur there.} In contrast, the top frequent items unique for LLM-generated data contain entries belonging to numeric and punctuation types, in other words things related to formal presentation of the text. We note also the words `fact' and `clear' as generic but persuasive, and as such perhaps typical for LLM language (Table 
\ref{tab:llm-lexentry-unique}).

\begin{table}[h!]
    \centering
    \small
    \begin{tabular}{lrl}
       Lex.\ entry & occurr.\ & example \\
       \hline
            ellipsis & 202 & `She was 86...' \\
            and\_or\_conj & 156 & `SF/SPCA' \\
            like\_comp & 125 &`It looks like the case...'\\
            num\_ne & 119 & `28th of July, 1966' \\
            square\_brack & 117 & `...using [the law]'\\
            time\_ne & 100 & `January 31st, 2019 5:34 pm' \\
            please\_root & 100 & `Please write to corrections'\\
            be\_nv\_is\_cx\_3 & 96 & `That's why we did it.'\\
            then\_adv & 82 & `by/since then'\\
            fact\_n2 & 81 & `The fact that...'\\
            wasn't & 81 & `It wasn't that loud' \\
            clear\_a2 & 70 & `It is not clear how.'\\
            OOV noun & 76 & `Anwar al-Awlaki' \\
            won't & 70 & `He won't care...'\\
            realize\_v2 & 69 &`I realized that...'\\
    \end{tabular}
    \caption{Frequent lexical entry usages unique for the LLaMa 65B-generated dataset}
    \label{tab:llm-lexentry-unique}
\end{table}


\section{Conclusion}
\label{sec:conclusion}
We present the first systematic comparison of LLM- and human-authored text through the lens of a formal grammatical theory (HPSG). We leverage the English Resource Grammar's explicit modeling of the principles of English syntax and lexicon, where detailed lexical types reflect the nuances of syntactic behavior of words. 

Comparing to the previous study by~\citet{munoz2024contrasting}, which used the same dataset but employed the UD syntactic framework, our analysis through the lens of formal syntactic theory confirms the validity of its conclusions even at a finer-grained level. It also offers greater detail on specific constructions that distinguish LLM-generated text from human-authored news. Our study also reaches novel conclusions on the same dataset by comparing individual human authors between each other as well as to LLMs.

We find that overall, LLMs tend to be more similar to each other along these grammatical dimensions than to humans. We show the importance of separating syntactic analysis from morphological, and that in the use of morphological rules, LLMs and humans are strikingly similar within the NYT genre. We find that human authors show greater variation between each other than a human-LLM pair; an LLM appears as an ``average'' human author. Further investigation of this syntactic and lexical flattening should be the subject of future papers, now that we have laid the groundwork of methodology, presented our analytical tools, and identified specific HPSG types to look into.  

Diversity indices show human-authored news as clearly distinct from all the LLMs (more diverse); however this is not so if we only look at lexical types. This opens up specific areas for future work.


We present some examples of constructions that occur more in human-authored than in LLM-generated news texts, and vice versa, confirming some but not other previous findings (such as the use of participial modifiers as more characteristic of LLM-generated text, which we do not confirm). Further experiments with various sampling techniques can provide further insight; in any case, using a resource such as the ERG is a way to ensure consistency and depth with respect to data analysis.

\section*{Limitations}
There are many methodological limitations related to work with LLM-generated text. An LLM will generate a different text every time, and a lot depends on the prompt, and our resources in terms of generation are limited. 
Otherwise the main limitation here is that we only look at one genre (NYT-style news). We do include other types of data and our analysis of the overall distribution reflects this; however in our discussion of specific examples we still focus on the NYT-style data. Another limitation is that we only have large HPSG grammars for a handful of languages, and indeed only the ERG is big enough to cover 94\% of news text, limiting the utility of our approach in comparisons of text in other languages. This is why our study is only about English.

\section*{Acknowledgments}
We thank Ann Copestake and Emily M.\ Bender, and more generally the DELPH-IN community for the useful discussion related to the paper.

We acknowledge grants SCANNER-UDC (PID2020-113230RB-C21) funded by MICIU/AEI/10.13039/501100011033; GAP (PID2022-139308OA-I00) funded by MICIU/AEI/10.13039/501100011033/ and ERDF, EU; LATCHING (PID2023-147129OB-C21) funded by MICIU/AEI/10.13039/501100011033 and ERDF, EU; and TSI-100925-2023-1 funded by Ministry for Digital Transformation and Civil Service and ``NextGenerationEU'' PRTR; as well as funding by Xunta de Galicia (ED431C 2024/02). 
CITIC, as a center accredited for excellence within the Galician University System and a member of the CIGUS Network, receives subsidies from the Department of Education, Science, Universities, and Vocational Training of the Xunta de Galicia. Additionally, it is co-financed by the EU through the FEDER Galicia 2021-27 operational program (Ref. ED431G 2023/01).

We have used ChatGPT for minor copy-editing (e.g.\ thesaurus suggestions) and for visualization ideas. We have used GitHub copilot for code autocompletion. 
\bibliography{llm-syntax}

\appendix

\section*{Appendices}
\section{English Resource Grammar types}
\label{sec:appendix:erg}
Table~\ref{tab:constructions} shows the construction types appearing in Figure \ref{fig:frequent-constr} with an expanded name and an example.   This is a slightly modified  version of the English Resource Grammar documentation.

\begin{table*}[ht]
\centering
\begin{tabular}{|l|l|l|}
\hline
\textbf{Type Name} & \textbf{Definition} & \textbf{Example} \\
\hline
\textbf{sb-hd\_mc\_c} & Head+subject, main clause & C arrived. \\ 
\textbf{sb-hd\_nmc\_c} & Hd+subject, embedded clause, subj has no gap & B thought [C arrived]. \\ 
\textbf{hd-cmp\_u\_c} & Hd+complement & B [hired C]. \\ 
\textbf{hd\_optcmp\_c} & Head discharges optional complement & B [ate] already. \\ 
\textbf{hdn\_optcmp\_c} & NomHd discharges opt complement & The [picture] appeared. \\ 
\textbf{mrk-nh\_evnt\_c} & Marker + event-based complement & B sang [and danced.] \\ 
\textbf{mrk-nh\_cl\_c} & Marker + clause & B sang [and C danced.] \\ 
\textbf{mrk-nh\_nom\_c} & Marker + NP & Cats [and some dogs] ran. \\ 
\textbf{mrk-nh\_n\_c} & Marker + N-bar & Every cat [and dog] ran. \\ 
\textbf{hd\_xcmp\_c} & Head extracts compl (to SLASH) & Who does B [admire] now? \\ 
\textbf{hd\_xsb-fin\_c} & Extract subject from finite hd & Who do you think [went?] \\ 
\textbf{sp-hd\_n\_c} & Hd+specifier, nonhd = sem hd & [Every cat] slept. \\ 
\textbf{sp-hd\_hc\_c} & Hd+specifier, hd = sem hd & The [very old] cat slept. \\ 
\textbf{aj-hd\_scp\_c} & Hd+preceding scopal adjunct & Probably B won. \\ 
\textbf{aj-hd\_scp-xp\_c} & Hd+prec.scop.adj, VP head & B [probably won]. \\ 
\textbf{hd-aj\_scp\_c} & Hd+following scopal adjunct & B wins if C loses. \\ 
\textbf{aj-hdn\_norm\_c} & Nominal head + preceding adjnct & The [big cat] slept. \\ 
\textbf{aj-hdn\_adjn\_c} & NomHd+prec.adj, hd pre-modified & The [big old cat] slept. \\ 
\textbf{aj-hd\_int\_c} & Hd+prec.intersective adjunct & B [quickly left]. \\ 
\textbf{hdn-aj\_rc\_c} & NomHd+following relative clause & The [cat we chased] ran. \\ 
\textbf{hdn-aj\_rc-pr\_c} & NomHd+foll.rel.cl, paired pnct & A [cat, which ran,] fell. \\ 
\textbf{hdn-aj\_redrel\_c} & NomHd+foll.predicative phrase & A [cat in a tree] fell. \\ 
\textbf{hd-aj\_int-unsl\_c} & Hd+foll.int.adjct, no gap & B [left quietly]. \\ 
\textbf{hd\_xaj-int-vp\_c} & Extract int.adjunct from VP & Here we [stand.] \\ 
\textbf{vp\_rc-redrel\_c} & Rel.cl. from predicative VP & Dogs [chasing cats] bark. \\ 
\textbf{hdn\_bnp\_c} & Bare noun phrase (no determiner) & [Cats] sleep. \\ 
\textbf{hdn\_bnp-pn\_c} & Bare NP from proper name & [Browne] arrived. \\ 
\textbf{hdn\_bnp-num\_c} & Bare NP from number & [42] is even. \\ 
\textbf{hdn\_bnp-qnt\_c} & NP from already-quantified dtr & [Some in Paris] slept. \\ 
\textbf{hdn\_bnp-vger\_c} & NP from verbal gerund & Hiring them was easy. \\ 
\textbf{np-hdn\_cpd\_c} & Compound from proper-name+noun & The [IBM report] arrived. \\ 
\textbf{np-hdn\_ttl-cpd\_c} & Compound from title+proper-name & [Professor Browne] left. \\ 
\textbf{np-hdn\_nme-cpd\_c} & Compound from two proper names & [Pat Browne] left. \\ 
\textbf{n-hdn\_cpd\_c} & Compound from two normal nouns & The [guard dog] barked. \\ 
\textbf{np\_adv\_c} & Modifier phrase from NP & B arrived [this week.] \\ 
\textbf{hdn\_np-num\_c} & NP from number & [700 billion] is too much. \\ 
\textbf{flr-hd\_nwh\_c} & Filler-head, non-wh filler & Kim, we should hire. \\ 
\textbf{flr-hd\_wh-nmc-fin\_c} & Fill-head, wh, fin hd, embed cl & B wondered [who won.] \\ 
\textbf{flr-hd\_rel-fin\_c} & Fill-head, finite, relative cls, NP gap & people [who we admired] \\ 
\textbf{vp-vp\_crd-fin-t\_c} & Conjnd VP, fin, top & B [sees C and chases D.] \\ 
\textbf{cl-cl\_crd-t\_c} & Conjoined clauses, non-int, top & B sang and C danced. \\ 
\textbf{np-np\_crd-t\_c} & Conjoined noun phrases, top & [The cat and the dog] ran. \\ 
\textbf{num-n\_mnp\_c} & Measure NP from number+noun & A [two liter] jar broke. \\ 
\textbf{cl\_np-wh\_c} & NP from WH clause & [What he saw] scared him. \\ 
\textbf{vp\_np-ger\_c} & NP from verbal gerund & Winning money [pleased C.] \\ 
\textbf{num\_det\_c} & Determiner from number & [Ten cats] slept. \\ 
\textbf{cl\_cnj-frg\_c} & Fragment clause with conjunctn & And Kim stayed. \\ 
\textbf{hd-pct\_c} & Head + punctuation token & B [arrived -] C left. \\ 
\textbf{hd-pct\_nobrk\_c} & Punctuation unrelated to bracketing &  \\ 
\textbf{pct-hd\_c} & Punctuation token + head & B arrived (today) \\ 

\hline
\end{tabular}
\caption{Construction types and examples.}\label{tab:constructions}
\end{table*}

\section{Cosine similarities}
\label{appendix:b}

Tables \ref{tab:cosine-syntax-only}-\ref{tab:cosine-lexrule-only} present the data underlying Figures \ref{fig:pca1}-\ref{fig:pca3} in \S\ref{sec:results}.

\begin{table}[h!]
\small
\begin{tabular}{lll}
\textbf{Model 1} & \textbf{Model 2} & \textbf{Cos} \\
\hline
llama\_30 & llama\_65 & 0.9999 \\
llama\_07 & llama\_13 & 0.9999 \\
llama\_07 & mistral\_07 & 0.9999 \\
llama\_13 & llama\_65 & 0.9998 \\
llama\_07 & llama\_65 & 0.9998 \\
llama\_13 & mistral\_07 & 0.9998 \\
llama\_13 & llama\_30 & 0.9998 \\
llama\_07 & llama\_30 & 0.9997 \\
llama\_65 & mistral\_07 & 0.9996 \\
llama\_30 & mistral\_07 & 0.9996 \\
falcon\_07 & llama\_30 & 0.9976 \\
falcon\_07 & mistral\_07 & 0.9972 \\
falcon\_07 & llama\_65 & 0.9972 \\
falcon\_07 & llama\_07 & 0.9966 \\
falcon\_07 & llama\_13 & 0.9966 \\
llama\_30 & \textbf{original NYT} & 0.9965 \\
llama\_65 & \textbf{original NYT} & 0.9964 \\
falcon\_07 & \textbf{original NYT} & 0.9958 \\
llama\_07 & \textbf{original NYT} & 0.9955 \\
mistral\_07 & \textbf{original NYT} & 0.9950 \\
llama\_13 & \textbf{original NYT} & 0.9950 \\
wsj & \textbf{original NYT} & 0.9949 \\
llama\_65 & wsj & 0.9908 \\
llama\_30 & wsj & 0.9907 \\
wikipedia & wsj & 0.9900 \\
llama\_07 & wsj & 0.9899 \\
mistral\_07 & wsj & 0.9894 \\
llama\_13 & wsj & 0.9891 \\
falcon\_07 & wsj & 0.9881 \\
wikipedia & \textbf{original NYT} & 0.9833 \\
llama\_65 & wikipedia & 0.9768 \\
llama\_07 & wikipedia & 0.9765 \\
llama\_30 & wikipedia & 0.9764 \\
mistral\_07 & wikipedia & 0.9763 \\
llama\_13 & wikipedia & 0.9745 \\
falcon\_07 & wikipedia & 0.9738 \\
\end{tabular}
\caption{Cosine similarity between LLM-generated and human-authored (\textit{original NYT}) datasets; only syntactic constructions included.}
\label{tab:cosine-syntax-only}
\end{table}

\begin{table}[h!]
\small
\begin{tabular}{lll}
\textbf{Model 1} & \textbf{Model 2} & \textbf{Cos} \\
\hline
llama\_30 & llama\_65 & 0.9999 \\
llama\_13 & llama\_65 & 0.9999 \\
llama\_07 & llama\_13 & 0.9999 \\
llama\_13 & llama\_30 & 0.9999 \\
llama\_07 & llama\_65 & 0.9998 \\
llama\_07 & llama\_30 & 0.9998 \\
llama\_07 & mistral\_07 & 0.9997 \\
llama\_13 & mistral\_07 & 0.9996 \\
llama\_30 & mistral\_07 & 0.9995 \\
llama\_65 & mistral\_07 & 0.9995 \\
falcon\_07 & llama\_30 & 0.9984 \\
falcon\_07 & llama\_13 & 0.9982 \\
falcon\_07 & llama\_65 & 0.9980 \\
falcon\_07 & llama\_07 & 0.9978 \\
falcon\_07 & mistral\_07 & 0.9977 \\
llama\_30 & \textbf{original NYT} & 0.9976 \\
llama\_65 & \textbf{original NYT} & 0.9975 \\
llama\_07 & \textbf{original NYT} & 0.9969 \\
llama\_13 & \textbf{original NYT} & 0.9968 \\
mistral\_07 & \textbf{original NYT} & 0.9965 \\
falcon\_07 & \textbf{original NYT} & 0.9956 \\
wsj & \textbf{original NYT} & 0.9922 \\
llama\_07 & wsj & 0.9909 \\
llama\_13 & wsj & 0.9908 \\
llama\_65 & wsj & 0.9906 \\
mistral\_07 & wsj & 0.9906 \\
llama\_30 & wsj & 0.9897 \\
falcon\_07 & wsj & 0.9837 \\
wikipedia & wsj & 0.9724 \\
wikipedia & \textbf{original NYT} & 0.9600 \\
llama\_07 & wikipedia & 0.9579 \\
mistral\_07 & wikipedia & 0.9579 \\
llama\_65 & wikipedia & 0.9570 \\
llama\_30 & wikipedia & 0.9565 \\
llama\_13 & wikipedia & 0.9559 \\
falcon\_07 & wikipedia & 0.9506 \\
\end{tabular}
\caption{Cosine similarity between LLM-generated and human-authored (\textit{original NYT}) datasets; only lexical type constructions included.}
\label{tab:cosine-lextype-only}
\end{table}

\begin{table}[h!]
\small
\begin{tabular}{lll}
\textbf{Model 1} & \textbf{Model 2} & \textbf{Cos} \\
\hline
llama\_13 & llama\_65 & 0.9999 \\
llama\_07 & llama\_65 & 0.9999 \\
llama\_30 & llama\_65 & 0.9999 \\
llama\_07 & llama\_13 & 0.9999 \\
llama\_13 & llama\_30 & 0.9998 \\
llama\_07 & mistral\_07 & 0.9998 \\
llama\_13 & mistral\_07 & 0.9997 \\
llama\_07 & llama\_30 & 0.9996 \\
llama\_65 & mistral\_07 & 0.9996 \\
llama\_30 & mistral\_07 & 0.9993 \\
llama\_65 & \textbf{original NYT} & 0.9990 \\
llama\_07 & \textbf{original NYT} & 0.9989 \\
llama\_30 & \textbf{original NYT} & 0.9989 \\
llama\_13 & \textbf{original NYT} & 0.9987 \\
mistral\_07 & \textbf{original NYT} & 0.9985 \\
falcon\_07 & llama\_30 & 0.9983 \\
falcon\_07 & llama\_13 & 0.9976 \\
falcon\_07 & llama\_65 & 0.9975 \\
falcon\_07 & llama\_07 & 0.9970 \\
falcon\_07 & mistral\_07 & 0.9966 \\
falcon\_07 & \textbf{original NYT} & 0.9962 \\
wsj & \textbf{original NYT} & 0.9932 \\
llama\_07 & wsj & 0.9923 \\
mistral\_07 & wsj & 0.9923 \\
llama\_65 & wsj & 0.9913 \\
llama\_13 & wsj & 0.9908 \\
llama\_30 & wsj & 0.9899 \\
falcon\_07 & wsj & 0.9822 \\
wikipedia & wsj & 0.9666 \\
mistral\_07 & wikipedia & 0.9476 \\
wikipedia & \textbf{original NYT} & 0.9474 \\
llama\_07 & wikipedia & 0.9464 \\
llama\_65 & wikipedia & 0.9427 \\
llama\_13 & wikipedia & 0.9418 \\
llama\_30 & wikipedia & 0.9393 \\
falcon\_07 & wikipedia & 0.9305 \\
\end{tabular}
\caption{Cosine similarity between LLM-generated and human-authored (\textit{original NYT}) datasets; only lexical rule constructions included.}
\label{tab:cosine-lexrule-only}
\end{table}

\section{Diversity Measures}
\label{appendix:c}

\begin{figure*}
  \begin{subfigure}[t]{0.5\textwidth}
        \centering
        \includegraphics[width=\columnwidth]{llm-erg-Constructions-Shannon}
      \end{subfigure}%
      ~ 
  \begin{subfigure}[t]{0.5\textwidth}
        \centering
        \includegraphics[width=\columnwidth]{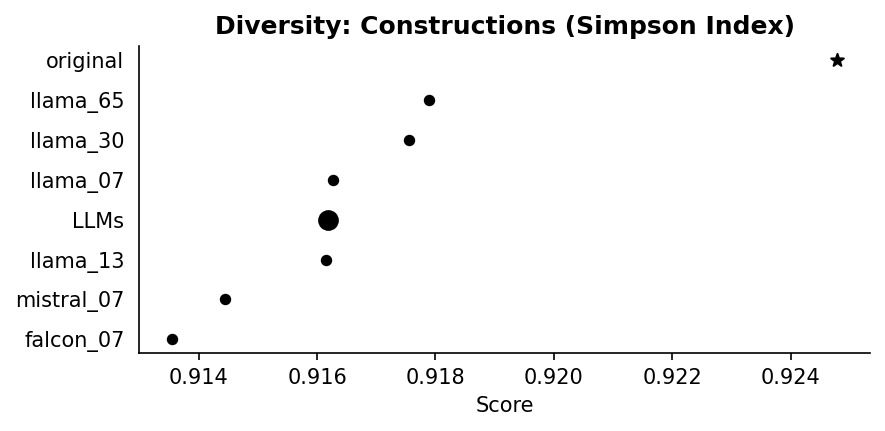}
      \end{subfigure}%
\\
      \begin{subfigure}[t]{0.5\textwidth}
        \centering
        \includegraphics[width=\columnwidth]{llm-erg-Lexical_Types-Shannon}
      \end{subfigure}
    ~ 
    \begin{subfigure}[t]{0.5\textwidth}
        \centering
        \includegraphics[width=\columnwidth]{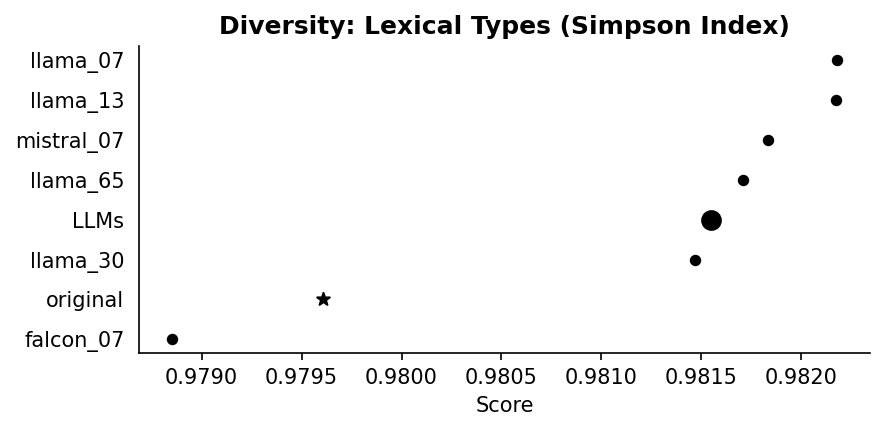}
      \end{subfigure}
      \\ 
    \begin{subfigure}[t]{0.5\textwidth}
        \centering
        \includegraphics[width=\columnwidth]{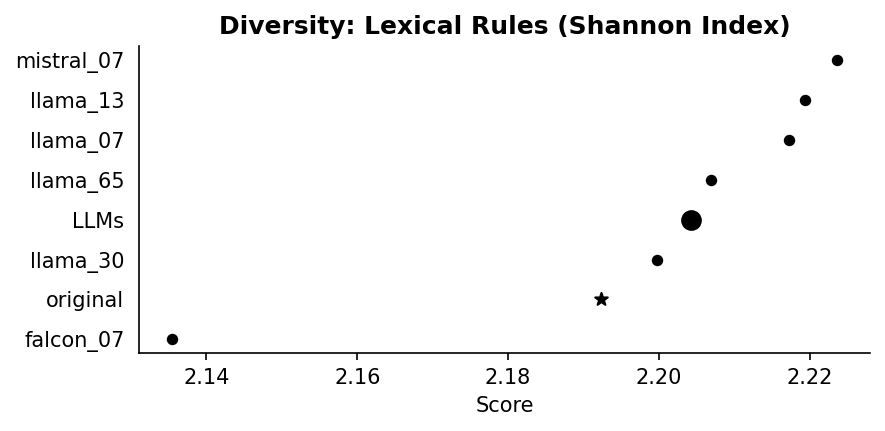}
      \end{subfigure}
 ~ 
    \begin{subfigure}[t]{0.5\textwidth}
        \centering
        \includegraphics[width=\columnwidth]{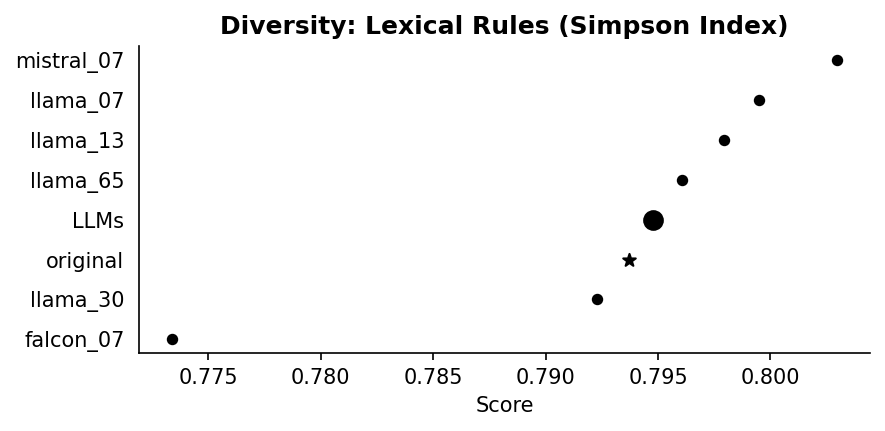}
    \end{subfigure}
      
    \caption{Diversity (Shannon and Simpson Indices)} \label{fig:shannon-simpson}
   \end{figure*}

Figure~\ref{fig:shannon-simpson} shows the diversities of constructions, lexical types and lexical rules measured with both the Shannon Index (on the left) and Simpson Index (on the right), as discussed in Section~\ref{sec:diversity}.   Scores for the original human-generated sentences are shown with a star ($\star$), LLMs with a dot ($\bullet$) and the combined LLMs with a larger dot ({\LARGE$\bullet$}) 

We measured the significance of the difference between the original human-generated sentences and combined LLM sentences using a permutation test, sampled 10,000 times.  All combinations had an observed p-value of less than 0.01, except for the Lexical Rules measured with the Simpson Index (which is less sensitive to outliers), with p = 0.13. 



\section{Mann-Whitney U-test}
\label{sec:appendix-d}
In this Appendix, we report the HPSG types for which the difference in relative frequency comes out as statistically significant (p $\leq$ 0.05; Tables \ref{tab:p-syntax}-\ref{tab:p-lexrules}). However, when we apply the FDR correction, none of these p-values remain below the 0.05 threshold. The definitions and examples for all of these HPSG types can be found in the English Resource Grammar files.\footnote{\url{https://github.com/delph-in/erg/releases/tag/2025}} The examples of where these types come up in the NYT corpus can be found in the data associated with this paper.\footnote{\url{https://github.com/olzama/llm-syntax/releases/tag/1.0.0}}

\onecolumn
\begin{longtable}{ll}
\caption{Mann-Whitney U-test (p $\leq$ 0.05) — Syntactic constructions} 
\label{tab:p-syntax}
\endfirsthead
\textbf{Frequent} & \\ \hline
aj-hd\_int\_c & 0.0238 \\
aj-hdn\_adjn\_c & 0.0238 \\
aj-hdn\_norm\_c & 0.0238 \\
cl-cl\_crd-t\_c & 0.0238 \\
cl\_cnj-frg\_c & 0.0476 \\
cl\_np-wh\_c & 0.0476 \\
flr-hd\_rel-fin\_c & 0.0238 \\
flr-hd\_wh-nmc-fin\_c & 0.0238 \\
hd-aj\_scp-pr\_c & 0.0238 \\
hd-aj\_vmod\_c & 0.0238 \\
hd-cmp\_u\_c & 0.0238 \\
hd-pct\_nobrk\_c & 0.0238 \\
hd\_xsb-fin\_c & 0.0238 \\
hdn\_bnp-qnt\_c & 0.0238 \\
hdn\_bnp\_c & 0.0238 \\
mrk-nh\_cl\_c & 0.0238 \\
mrk-nh\_n\_c & 0.0238 \\
mrk-nh\_nom\_c & 0.0476 \\
n-hdn\_cpd\_c & 0.0238 \\
np-np\_crd-t\_c & 0.0238 \\
num\_det\_c & 0.0476 \\
sb-hd\_mc\_c & 0.0238 \\
vp\_rc-redrel\_c & 0.0238 \\
vp\_sbrd-prd-prp\_c & 0.0238 \\
\addlinespace[0.5em]
\textbf{Infrequent} & \\ \hline
aj-hd\_int-inv\_c & 0.0238 \\
aj-hdn\_crd-cma\_c & 0.0238 \\
cl-cl\_crd-int-t\_c & 0.0238 \\
cl-np\_runon\_c & 0.0238 \\
cl\_rc-inf-modgap\_c & 0.0476 \\
cl\_rc-inf-nwh\_c & 0.0476 \\
flr-hd\_nwh-nmc\_c & 0.0238 \\
flr-hd\_wh-mc\_c & 0.0476 \\
flr-hd\_wh-nmc-inf\_c & 0.0238 \\
hd-aj\_cmod-s\_c & 0.0476 \\
hd-aj\_vmod-s\_c & 0.0238 \\
hd-hd\_rnr-nb\_c & 0.0476 \\
hd-hd\_rnr-nv\_c & 0.0476 \\
hd-hd\_rnr\_c & 0.0238 \\
hdn-aj\_rc-asym\_c & 0.0238 \\
hdn-aj\_rc-propr\_c & 0.0238 \\
hdn-aj\_redrel-asym\_c & 0.0238 \\
hdn-aj\_redrel-pr\_c & 0.0238 \\
hdn-np\_app-dx\_c & 0.0238 \\
hdn-np\_app-mnp\_c & 0.0238 \\
j-j\_crd-att-t\_c & 0.0476 \\
j-n\_crd-m\_c & 0.0476 \\
j-n\_crd-t\_c & 0.0238 \\
j\_n-ed\_c & 0.0476 \\
mrk-nh\_atom\_c & 0.0238 \\
n-hdn\_cpd-pl-mnp\_c & 0.0431 \\
n-hdn\_cpd-pl\_c & 0.0238 \\
n-j\_j-cpd\_c & 0.0238 \\
n-j\_j-t-cpd\_c & 0.0238 \\
n-n\_crd-asym-t\_c & 0.0238 \\
n-n\_crd-div-t\_c & 0.0238 \\
n-n\_crd-im\_c & 0.0238 \\
n-n\_num-seq\_c & 0.0275 \\
n-v\_j-cpd\_c & 0.0238 \\
np-np\_crd-im\_c & 0.0238 \\
np-np\_crd-nc-m\_c & 0.0238 \\
np\_indef-adv\_c & 0.0476 \\
np\_nb-pr-frg\_c & 0.0238 \\
num\_prt-det-nc\_c & 0.0238 \\
num\_prt-of\_c & 0.0476 \\
pp-pp\_crd-im\_c & 0.0476 \\
pp-pp\_crd-t\_c & 0.0238 \\
r\_cl-frg\_c & 0.0476 \\
sb-hd\_q\_c & 0.0238 \\
vp\_sbrd-prd-pas\_c & 0.0238 \\
vp\_sbrd-pre-lx\_c & 0.0238 \\
vp\_sbrd-pre\_c & 0.0238 \\
\end{longtable}

\begin{longtable}{ll}
\caption{Mann-Whitney U-test (p $\leq$ 0.05) — Lexical types}
\label{tab:p-lextypes}
\endfirsthead
\textbf{Frequent} & \\ \hline
aj\_-\_i-att\_le & 0.0238 \\
aj\_-\_i-ord-one\_le & 0.0238 \\
aj\_pp\_i-er\_le & 0.0238 \\
aj\_vp\_i-seq\_le & 0.0238 \\
av\_-\_dg-cmp-so\_le & 0.0238 \\
av\_-\_dg-jo\_le & 0.0238 \\
av\_-\_dg-sup\_le & 0.0238 \\
av\_-\_i-vp-pr\_le & 0.0476 \\
av\_-\_i-vp\_le & 0.0238 \\
c\_xp\_but\_le & 0.0238 \\
cm\_np-vp\_that\_le & 0.0238 \\
cm\_vp\_to\_le & 0.0238 \\
d\_-\_poss-my\_le & 0.0238 \\
d\_-\_poss-our\_le & 0.0476 \\
d\_-\_poss-their\_le & 0.0476 \\
d\_-\_poss-your\_le & 0.0476 \\
n\_-\_ad-pl\_le & 0.0476 \\
n\_-\_c-ed-ns\_le & 0.0238 \\
n\_-\_c-nocnh-cap\_le & 0.0238 \\
n\_-\_c-ns\_le & 0.0238 \\
n\_-\_c-time\_le & 0.0238 \\
n\_-\_m-time\_le & 0.0476 \\
n\_-\_m\_le & 0.0238 \\
n\_-\_mc\_le & 0.0238 \\
n\_-\_pn-sg\_le & 0.0238 \\
n\_-\_pn-yoc-gen\_le & 0.0238 \\
n\_-\_pr-dei-sg\_le & 0.0476 \\
n\_-\_pr-he\_le & 0.0238 \\
n\_-\_pr-i\_le & 0.0275 \\
n\_-\_pr-it-x\_le & 0.0238 \\
n\_-\_pr-it\_le & 0.0238 \\
n\_-\_pr-me\_le & 0.0238 \\
n\_-\_pr-rel-who\_le & 0.0238 \\
n\_-\_pr-she\_le & 0.0238 \\
n\_-\_pr-them\_le & 0.0476 \\
n\_-\_pr-they\_le & 0.0238 \\
n\_-\_pr-we\_le & 0.0238 \\
n\_-\_pr-wh\_le & 0.0476 \\
n\_-\_pr-you\_le & 0.0238 \\
n\_-\_pr\_le & 0.0476 \\
n\_pp\_c-ns\_le & 0.0238 \\
n\_pp\_c-nsnc-of\_le & 0.0238 \\
n\_pp\_c-pl\_le & 0.0238 \\
n\_pp\_m\_le & 0.0476 \\
n\_vp\_c\_le & 0.0238 \\
p\_cp\_s\_le & 0.0476 \\
p\_np\_i-ngap\_le & 0.0238 \\
p\_np\_i-nm-poss\_le & 0.0476 \\
p\_np\_ptcl\_le & 0.0476 \\
pp\_-\_i-wh\_le & 0.0238 \\
pt\_-\_bang\_le & 0.0238 \\
pt\_-\_comma-informal\_le & 0.0238 \\
pt\_-\_hyphn-rgt\_le & 0.0238 \\
pt\_-\_period\_le & 0.0238 \\
v\_cp\_fin-inf-q\_le & 0.0238 \\
v\_cp\_prop\_le & 0.0238 \\
v\_np-cp\_fin-inf\_le & 0.0238 \\
v\_np-pp\_prop\_le & 0.0238 \\
v\_np-vp\_bse\_le & 0.0238 \\
v\_np-vp\_oeq\_le & 0.0238 \\
v\_np\_be\_le & 0.0238 \\
v\_np\_is-cx\_le & 0.0238 \\
v\_np\_le & 0.0238 \\
v\_np\_poss\_le & 0.0238 \\
v\_np\_was\_le & 0.0238 \\
v\_pp*-pp*\_le & 0.0238 \\
v\_prd\_are-cx\_le & 0.0238 \\
v\_prd\_been\_le & 0.0238 \\
v\_prd\_being\_le & 0.0238 \\
v\_prd\_is-cx\_le & 0.0238 \\
v\_prd\_was\_le & 0.0238 \\
v\_prd\_wre\_le & 0.0238 \\
v\_vp\_has\_le & 0.0238 \\
v\_vp\_have-f\_le & 0.0238 \\
v\_vp\_seq\_le & 0.0238 \\
v\_vp\_ssr\_le & 0.0238 \\
\addlinespace[0.5em]
\textbf{Infrequent} & \\ \hline
aj\_-\_i-att-er\_le & 0.0091 \\
aj\_-\_i-one-nmd\_le & 0.0339 \\
av\_-\_i-unk\_le & 0.0091 \\
n\_-\_c-meas\_le & 0.0091 \\
n\_-\_c-min\_le & 0.0091 \\
n\_-\_m-hldy\_le & 0.0091 \\
n\_-\_pn-abb\_le & 0.0339 \\
n\_-\_pn-unk\_le & 0.0091 \\
n\_-\_pr-her\_le & 0.0091 \\
n\_pp\_c-dir\_le & 0.0091 \\
pp\_-\_i-po-tm\_le & 0.0091 \\
\end{longtable}

\begin{longtable}{ll}
\caption{Mann-Whitney U-test (p $\leq$ 0.05) — Lexical rules} 
\label{tab:p-lexrules}
\endfirsthead
\textbf{Frequent} & \\ \hline
n\_det-mnth\_dlr & 0.0238 \\
n\_pl-irreg\_olr & 0.0238 \\
n\_pl\_olr & 0.0238 \\
v\_aux-cx-noinv\_dlr & 0.0238 \\
v\_j-nb-pas-tr\_dlr & 0.0238 \\
v\_n3s-bse\_ilr & 0.0238 \\
v\_nger-tr\_dlr & 0.0238 \\
v\_psp\_olr & 0.0238 \\
\addlinespace[0.5em]
\textbf{Infrequent} & \\ \hline
det\_prt-of-agr\_dlr & 0.0476 \\
j\_enough-wc-nogap\_dlr & 0.0476 \\
j\_j-non\_dlr & 0.0238 \\
j\_j-un\_dlr & 0.0476 \\
j\_tough-compar\_dlr & 0.0238 \\
n\_n-hour\_dlr & 0.0238 \\
v\_aux-ell-ref\_dlr & 0.0476 \\
v\_aux-ell-xpl\_dlr & 0.0476 \\
v\_aux-sb-inv\_dlr & 0.0476 \\
v\_aux-tag\_dlr & 0.0238 \\
v\_j-nb-intr\_dlr & 0.0238 \\
v\_j-nb-pas-ptcl\_dlr & 0.0238 \\
v\_j-nme-tr\_dlr & 0.0238 \\
v\_v-pre\_dlr & 0.0476 \\
v\_v-re\_dlr & 0.0238 \\
v\_v-un\_dlr & 0.0238 \\
w\_mwe-3-wb\_dlr & 0.0219 \\
w\_mwe-wb\_dlr & 0.0476 \\
\end{longtable}

\end{document}